\documentclass[letterpaper, 10 pt, conference]{ieeeconf}  
\usepackage{algorithm}
\usepackage{algpseudocode}
\usepackage{graphicx}
\usepackage{float}
\usepackage{amsmath}
\usepackage{soul}
\usepackage{color}

\IEEEoverridecommandlockouts                              

\title{\LARGE \bf
ARO: Large Language Model Supervised Robotics Text2Skill Autonomous Learning
}

\author{Yiwen Chen$^{1}$, Yuyao Ye$^{2}$, Ziyi Chen$^{2}$, Chuheng Zhang$^{3}$, Marcelo H. Ang$^{4}$ 
\thanks{*This work was not supported by any organization}
\thanks{$^{1}$Yiwen Chen, National University of Singapore
        {\tt\small yiwen.chen@u.nus.edu}}%
\thanks{$^{2}$Yuyao Ye, National University of Singapore
        {\tt\small e1192851@u.nus.edu}}%
\thanks{$^{2}$Ziyi Chen, National University of Singapore
        {\tt\small e1192778@u.nus.edu}}%
\thanks{$^{3}$Chuheng Zhang, Microsoft Research
        {\tt\small zhangchuheng123@live.com}}%
\thanks{$^{4}$Marcelo H. Ang, National University of Singapore
        {\tt\small mpeangh@nus.edu.sg}}%
}

\begin{document}

\maketitle
\thispagestyle{empty}
\pagestyle{empty}

\begin{abstract}

Robotics learning highly relies on human expertise and efforts, such as demonstrations, design of reward functions in reinforcement learning, performance evaluation using human feedback, etc. However, reliance on human assistance can lead to expensive learning costs and make skill learning difficult to scale. In this work, we introduce the Large Language Model Supervised Robotics Text2Skill Autonomous Learning (ARO) framework, which aims to replace human participation in the robot skill learning process with large-scale language models that incorporate reward function design and performance evaluation. We provide evidence that our approach enables fully autonomous robot skill learning, capable of completing partial tasks without human intervention. Furthermore, we also analyze the limitations of this approach in task understanding and optimization stability. 

\end{abstract}

\section{INTRODUCTION}
In the pursuit of autonomous robotics, a paradigm shift is being observed with the integration of Large Language Models (LLMs) into the learning process. The current landscape reveals an overt reliance on human experts to script and fine-tune robotic behaviors\cite{yu2023language}, a process that is both time-consuming and limited in scalability. Traditional control methods, with their rigid programming structures, do not dynamically adapt to diverse tasks without extensive human intervention. In contrast, while reinforcement learning (RL) presents a more flexible approach, it also relies heavily on human experience to design reward functions, evaluate results, and create interaction environments\cite{ng1999policy}. The rapid development of LLM has led to the expectation that the skill-learning process of robots can be optimized by combining LLM with RL, thus reducing the reliance on human experts.

The conundrum lies in the pre-LLM challenges: Despite the remarkable capabilities of LLMs in understanding and generating natural language, translating this into actionable tasks for robots remains an intricate hurdle. A plausible solution is the use of LLMs in the design of reward functions that not only accurately reflect desired behaviors, but also exhibit generalizability. This requires an iterative cycle of experimentation and testing by human experts to pinpoint optimal directions. Imagine a paradigm in which robots autonomously learn, refine skills, and adapt to diverse tasks without human intervention, marking a significant leap in autonomous robotics. This framework suggests a transformative era of enhanced capability and versatility that redefines the potential of robotic autonomy.

Driven by this vision, our work introduces an LLM-supervised Robotics Text2Skill Autonomous Learning (ARO) framework.\footnote{https://github.com/yiwc/aro} By inputting straightforward natural language instructions, our model yields a corresponding RL neural network adept at manipulating a robot to execute actions that resonate with the given directives. Our methodology harnesses the interpretative and generative strengths of LLMs to interpret task descriptions and generate a suitable reward function code. This code is then employed to train a specific RL agent. Subsequently, the LLM designs an evaluation function to assess the performance of the robot under the agent's control, producing performance data. The LLM then analyzes the data to determine if the performance matches the initial requirements. In cases where performance falls short, the LLM provides suggestions for improvement, which are fed back as prompts to iteratively refine the reward function code. 

Our approach implements a robot learning model that is completely free from the reliance on human experts. The entire model can be automated to achieve near-end-to-end results. This approach significantly alleviates the need for manually coded reward functions, and its adaptability to a wide spectrum of tasks is enhanced by its independence from human expert evaluation criteria. Applying LLM to robots with embodied intelligence gives robots an initial grasp of "intelligence", and our approach is helping robots learn how to use their initial "intelligence" to solve more difficult problems. Rather than increasing the robot's "intelligence" - i.e., augmenting the capabilities of the LLM - the future also lies in enabling robots to learn to make better use of the LLM.

\section{Related Work}

\textbf{Reward Generation}. The evolution of reward generation methodologies, particularly through the integration of Large Language Models, represents a notable advancement in the field of robotics. Traditional approaches to the design of reward functions have often required significant manual tuning and expert knowledge to capture the nuances of desired behaviors within robotic tasks. However, recent developments have demonstrated the potential of LLMs to streamline this process. For instance, the work presented in \textit{Eureka} \cite{ma2023eureka} explores the capability of LLMs to generate complex reward functions that closely align with human reasoning, reducing the dependency on manual design processes. \textit{T2R} \cite{xietext2reward} proposes a strategy in which an LLM designs a reward function and humans provide feedback, subsequently training the RL agent. For the task of controlling a robot, two different prompts, zero-shot and few-shot, are given. After the reward is generated by GPT-4, the user provides feedback and then controls the robot to walk. Similarly, \textit{Language to Rewards for Robotic Skill Synthesis} \cite{yu2023language} demonstrates how natural language descriptions of tasks can be directly translated into actionable reward functions by LLM, bridging the gap between human conceptualization of tasks and robotic execution. These approaches highlight the significant impact of LLMs on automating and refining the reward generation process, allowing robots to understand and perform tasks with increased flexibility and efficiency.

\textbf{Grounding in Robotics}. The grounding of language in robotics has been an active area of research aimed at establishing connections between linguistic constructs and robotic perception and action. The work \textit{Do As I Can} \cite{ahn2022can} presents a novel approach to grounding leveraging the concept of advantages. This method allows robots to interpret tasks contextually, enhancing their ability to execute instructions that align with their physical capabilities. Another notable contribution is \textit{Reflexion} \cite{shinn2024reflexion}, which introduces verbal feedback as a mechanism for language grounding. The model demonstrates improved task execution by incorporating verbal cues into the learning process.

However, the primary challenge in this domain remains the complexity of mapping abstract language to concrete actions. While the proposed methods show promise, the ambiguity and diversity of natural language continue to pose difficulties for precise execution in varied environments.

\textbf{Learning through LLM}. The intersection of LLM and robotic learning represents a growing field of research. \textit{Self-Refined Large Language Model} \cite{song2023self} demonstrates how LLMs can autonomously generate reward functions, significantly reducing the need for human-designed rewards. \textit{RoboGen} \cite{wang2023robogen} exemplifies the use of generative models to create diverse training data, illustrating how LLMs can expand the scope of learning for robots beyond limited real-world experiences.

The integration of LLM into robotic learning has shown considerable potential, particularly in terms of data generation and reward shaping. However, the dependency on high-quality data and the generalizability of learned models across different tasks remain areas that require further investigation.

\textbf{Planning with Embodied AI}. Planning is a critical aspect of autonomous decision making in embodied AI. The research presented in \textit{PROGROMPT} \cite{singh2023progprompt} takes a pioneering step toward the use of LLMs for task planning. This work showcases the vast knowledge base of LLMs to formulate plans that are sensitive to the context and adaptable. Furthermore, \textit{PalM-E} \cite{driess2023palm} integrates multimodal data to improve robot planning capabilities, allowing more nuanced and situationally appropriate action plans.

Although these studies mark significant progress in robotic planning, challenges such as computational efficiency and alignment of plans with real-time environmental changes require further exploration. Moreover, ensuring that the plans are executable within the physical constraints of the robot's design requires ongoing refinement.

\section{Approach}

\subsection{Overview}

For reward function generation, adaptive effectiveness evaluation, and iterative optimization, ARO contains five main modules:

\begin{itemize}
    \item \textbf{Reward Function Generator(RFG) Module} generates the Reward Function (RF) code to use the reward function according to the environment in which it is written.
    \item \textbf{Training Module} uses the current environment to carry out reinforcement learning training based on the Soft Actor-Critical(SAC) model.
    \item \textbf{Evaluation Function Generator(EFG) Module} generates a custom Evaluation Function (EF) code and generates performance data.
    \item \textbf{Performance Evaluator(PE) Module} is used to evaluate performance data, find problems, and make suggestions.
    \item \textbf{Environment Evaluator(EE) Module} converts improvement suggestions and environment code into an environment description to help the agent better understand environmental information.
\end{itemize}

\begin{figure}[h]
    \centering
    \includegraphics[width=0.75\linewidth]{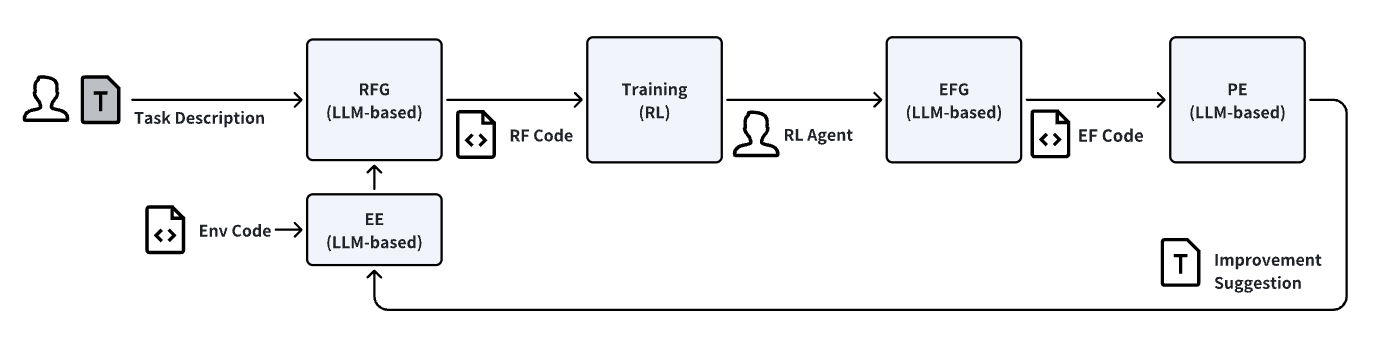}
    \caption{ARO's Framework}
    \label{fig:v1}
\end{figure}

The whole process is demonstrated by Algo \ref{a1} and Fig. \ref{fig:v1}.

\begin{algorithm}
\caption{ARO System Process with Iterative Training and Refinement} \label{a1}
\begin{algorithmic}[1] 
\Require Task Description $T$, Environment Code $E$

\State $T_{imp} \gets \{\}$ \Comment{Initialize Improvement Suggestions as empty}

\While{not ConvergenceCriteriaMet($Agent$)}
    \State $T_{EnvDesc} \gets $ $F_{EE}$($T_{Env}$,$T_{imp}$) \\
    \Comment{Generate Environment Explanation}
    \State  $T_{RF} \gets$ $F_{RFG}$($T_{task}, T_{Env}, T_{example}, T_{imp}$) \\ \Comment{Generate RF code using RFG Module}
    \State $\theta \gets$ Train($T_{RF}$, $T_{Env}$,)\\ \Comment{Train the RL agent using Training Module}
    \State $T_{Perf}\gets$ $F_{EFG}$($T_{task}, T_{Evv}, T_{example}, T_{imp}, \theta$)\\ \Comment{Generate Performance Data using EF Module}
    \State $T_{imp} \gets$ $F_{PE}$($T_{task},T_{Env},T_{example},T_{RF},T_{EF},T_{Perf}$) \\ \Comment{Generates the improvemen suggestions for others.}
\EndWhile
\State \textbf{Output:} $\theta$
\end{algorithmic}
\end{algorithm}

\textbf{Iteration}. Central to the ARO framework's training regimen is an iterative process that dynamically refines the reward function in response to evaluative feedback. The iteration commences with the EFG analyzing the trained model's performance against the task description. Feedback from this evaluation influences subsequent modifications to the RFG. 

This cyclical process iterates as follows:
\begin{itemize}
    \item The EFG module evaluates the current model's behavior and outputs performance metrics alongside qualitative assessments. Based on these evaluations, the ARO identifies discrepancies between current behavior and desired task outcomes.
    \item The RFG module ingests this feedback to adjust the reward function, in order to rectify the identified performance gaps. The updated reward function is then applied, and the model undergoes further training.
    \item The cycle repeats, with the EFG reassessing the model post-adjustment, ensuring continuous enhancement until the performance metrics align with predefined success criteria.
\end{itemize}

This iterative feedback loop is crucial for the development of agent behavior. Allows for autonomous refinement of actions, reduces the dependence on predefined reward structures, and allows the model to achieve optimal task execution with minimal human intervention. Convergence is achieved when the agent's performance meets the criteria set by the task description, indicating successful training and successful skill acquisition by the robot.

\subsection{RFG Module}

\begin{algorithm}
\caption{Reward Function Generator (RFG) Module} \label{a2}
\begin{algorithmic}[1] 
\Require Task Description $T_{task}$, Environment Code $T_{Env}$, Example Reward Function Code $T_{example}$, Improvement Suggestions $T_{imp}$ (if any)
\For{$N$ iterations}
    \State $T_{RF} \gets$ $F_{RFG}$($T_{task}, T_{Env}, T_{example}, T_{imp}$)
    \State $success, errors \gets$ TestRFCode($T_{RF}, T_{Env}$)
    \If{$success$}
        \State \textbf{break}
    \Else
        \State $T_{RF} \gets$ UpdateRFG($T_{error}, T_{RF}$)
    \EndIf
\EndFor
\State \textbf{Output:}  Executable Reward Function Code  $T_{RF}$
\end{algorithmic}
\end{algorithm}

To solve the reward function design problem, we propose the RFG Module, which can help solve the reward function design challenge in reinforcement learning training. Its process is demonstrated by the Algo \ref{a2}.

The process begins with the input of a description of a natural language task, such as \textit{ training a humanoid robot to walk like a human}. The RFG module utilizes GPT-4 to analyze the input and generate RF code suitable for training an RL agent. If the generated code is non-executable or results in errors, the feedback (including the code and error messages) is reinputted into the RFG module for further refinement. This iterative process continues until a viable RF code is produced.

\begin{itemize}
    \item \textbf{Module Input.} The task description, environment code, example code, improvement suggestions (if any), last generated RF code (if any) and error message (if any) are entered into the RFG module to construct a prompt for GPT-4 \cite{achiam2023gpt}. The prompt directs GPT-4 to generate a compliant RF code.
    \item \textbf{Iteration}. Generate a code script that applies the RF code to the environment using GPT-4 in the RFG module. And try to run the code, and if there is an error message, input the error message and the previously generated RF code feedback model. Iterate the code until it is ready to run.
    \item \textbf{Output}. The output is the result of a function that can be integrated into the reward function of the reinforcement learning lab environment.
\end{itemize}

\subsection{Training module}
In order to translate the knowledge in the reward function, into the agent's skills and actions, we propose the Training Module. The RF code is used to train an RL agent based on the Soft Actor-Critic (SAC) algorithm\cite{christodoulou2019soft}\cite{haarnoja2018soft}. The trained agent embodies a neural network capable of controlling the robot. 

\begin{itemize}
    \item \textbf{Module Input}. Simulator environment for simulating robot dynamics, physics, collisions, etc. The reward function code, generated and refined, is used to train the intelligent body, ensuring that the behavior of the intelligent body matches the task description.
    \item \textbf{Iteration}.  This process will take advantage of the SAC algorithm to train the agent to learn to control the robot to perform behaviors that match the description of the task.
    \item \textbf{Output}. Ultimately, the training process produces an RL agent that contains a neural network that is capable of making decisions based on the observed state of the environment to perform complex control tasks, such as robot motion control.
\end{itemize}

\subsection{EFG Module}

\begin{algorithm}
\caption{Evaluation Function Generator (EFG) Module} \label{a3}
\begin{algorithmic}[1]
\Require Task Description $T_{task}$, Environment Code $T_{Env}$, Example Evaluation Function Code $T_{example}$, Improvement Suggestions $T_{imp}$ (if any), Trained Agent, $\theta$

\For{$N$ iterations}
    \State $T_{EF} \gets$ $F_{EFG}$($T_{task}, T_{Env}, T_{example}, T_{imp}, \theta$)
    \State $success, errors \gets$ TestEFCode($T_{EF}, T_{Env}$)
    \If{$success$}
        \State \textbf{break}
    \EndIf
\EndFor

\State $T_{Perf} \gets$ EvaluateModel($T_{EF}, T_{env}, \theta$) \Comment{Runs EF code to get Performance Data}
\State \textbf{Output:} Executable Evaluation Function Code $T_{EF}$, Performance Data$T_{Perf}$
\end{algorithmic}
\end{algorithm}

Upon completion of the training phase, the model is evaluated using the EFG module, which generates the EF code to assess the performance of the reinforcement learning agent. To evaluate the performance of our trained models, we designed the EFG module, which can be used in GPT-4 to evaluate training results. The Algo \ref{a3} demonstrates the contents of the EFG module. The EF code follows a structured process that encapsulates the interaction with the GPT-4 model and generates scripts that meet specific requirements. The code was then validated and executed to generate performance data.

\begin{itemize}
    \item \textbf{Module Input.} The task description, environment code, example code, improvement suggestions (if any), last generated EF code (if any) and error message (if any) are entered into the EFG module to construct a prompt for the GPT-4. The prompt directs GPT-4 to generate a compliant EF code.
    \item \textbf{Iteration}. Create a prompt that summarizes the task description and any suggestions for improvement. And use the GPT-4 model to generate EF code tailored to evaluate model performance. Then validate the generated EF code to ensure that it can run. Execute the EF code to assess the model's ability, tracking metrics.
    \item \textbf{Output}. Performance data including metrics relevant to the behavior. The metrics are generated by module, and it is not pre-designed by human experts.
\end{itemize}

\subsection{PE Module}
To enable the optimization process of the model, we propose the PE module that autonomously analyzes performance data and provides feasible improvements. The module closes the gap between the current performance of the model and the target results of the task description.

The satisfaction of the performance data is determined by the $\textrm{IsSatisfied}$ function \ref{IsS}:

\begin{equation}
\textrm{IsSatisfied}(T_{Perf}) =
\begin{cases}
1, & \textrm{if performance is satisfactory} \\
0, & \textrm{otherwise}
\end{cases} \label{IsS}
\end{equation}

If performance is satisfactory ($\textrm{IsSatisfied}(T_{Perf}) = 1$), the training process is terminated. Otherwise, the PE module generates improvement suggestions $T_{imp}$ using the $F_{PE}$ function \ref{FPE}:

\begin{equation}
T_{imp} = F_{PE}(T_{task}, T_{Env}, T_{example}, T_{RF}, T_{EF}, T_{Perf}) \label{FPE}
\end{equation}

These suggestions for improvement provide feedback to the iterative learning process, guiding ARO to make further improvements to the output.

Integrating descriptive and evaluative modules into the PE module enables an automated and comprehensive evaluation process that improves the accuracy of performance feedback and reduces the need for laborious manual optimization practices.

\begin{itemize}
    \item \textbf{Module Input}. Initially, the module ingests content pertinent to task description, environment code, example code, RF code, EF code and performance data through their respective documentation.
    \item \textbf{Output}. By analyzing a complete set of inputs, the PE module systematically identifies performance deficiencies and autonomously generates feasible modification recommendations. These recommendations include adjustments to the reward function, refinements to the evaluation criteria, and improvements to the description of the environment.
\end{itemize}

\subsection{EE Module}
   
\begin{equation}
    T_{EnvDesc} = F_{EE}(T_{Env},T_{imp}) \label{a5}
\end{equation}

To solve the problem of too much information in the environment, which is difficult to process by the model, we design the EE module to use GPT-4 to extract the key information to help the model understand the environment better. Eq.\ref{a5} demonstrates the content of the EE module. The EE module is designed to automatically generate detailed information and explanations about a specific environment using the GPT-4 model.

\begin{itemize}
    \item \textbf{Module Input}. The improvement suggestions(if any) given by the PE module and the environment code are fed into the EE module as input.
    \item \textbf{Output}. Verified generated information, i.e., the environment explanation, is returned as Environment Description($T_{Envdesc}$).
\end{itemize}

\section{Experiment}
We completely evaluated ARO's performance in different contexts, from simple skills to complex actions, and verified the validity of the model. We tested three different sets of standing, forward walking, and backward walking tasks for the robot, and ARO was able to properly control the robot to perform the task description. In addition, we also used LLM to randomly generate 29 task requirements to test the response effect and quality of our model under complex and changeable prompt conditions. Without the need for human experts to provide advice and iteration, ARO achieves a complete reinforcement learning process without expert supervision and allows the robot to acquire new skills. Our experimental results are based on reliable data and are reproducible; the final model output results in a complete and runnable RL agent.

\textbf{Environment}. Our experimental environment is based on gym\cite{brockman2016openai}, SAC is implemented from Stable-baselines3\cite{raffin2019stable}, and GPT-4 APIs\cite{achiam2023gpt}. Our simulation environment uses three of the MuJoCo's from Gymnasium robots to realize randomly generated tasks. First, we use the MuJoCo humanoid, Hopper, and HalfCheetah robot dynamics models, which encompass the different forms of robots and include the main robot types. Second, we design environments to simulate real physics, allowing realistic task execution for standing still, forward motion, and backward motion. Integration with the gym interface allows for standardized task execution and performance metrics, facilitating seamless interaction between the SAC algorithms and the robot models. Our setup combines gym's environmental standardization, MuJoCo's simulations, SAC reinforcement learning algorithms, and ARO-v1's reward function generation and autonomous performance evaluation and learning iteration, offering a robust platform for studying autonomous robot skill learning.

\textbf{Metrics} We experimented with ARO to perform the 29 randomly generated tasks we asked and judged the strengths and weaknesses of the ARO model based on the success rate of reward generation and task completion.

\textbf{Reward Generation} Reward Generation method refers to the agent trained by the reward function designed without the iteration of ARO, as it is designed directly by RFG module without the modification of ARO iteration, representing the ability of RFG module to design the reward function.

\section{Results and Analysis}

\begin{table}[h]
    \centering
    \caption{Basic Experiment Results}
    \label{tab:ExperimentResults}
    \begin{tabular}{c|ccc}
        \hline
        Task / Robot & Humanoid & Hopper & HalfCheetah \\ \hline
        Stand Still & Successful & Successful & Successful \\
        Forward Motion & Successful & Successful & Successful \\
        Backward Motion & Successful & Successful & Successful \\ \hline
    \end{tabular}
\end{table}

\begin{table*}[h]
\centering
\caption{Comparison on Language to Action task environments} \label{tab: comparison}
\label{tab:Comparison}
\begin{tabular}{c|l|cc}
\hline
Task &Environment&Reward Generation & ARO Output \\ \hline
Combine jumping and forward movement&Halfcheetah&Successful& Successful\\
Execute quick turns while sprinting&Halfcheetah&Failed& Failed\\
Jump as high as possible from a stationary position&Halfcheetah&Successful& Failed\\
Maintain a low stance while moving fast forward&Halfcheetah& Successful&Successful\\
Reduce leg movements to simulate gliding for a distance&Halfcheetah& Failed&Failed\\
Simulate curling and stretching actions&Halfcheetah& Failed&Failed\\
 Sprint backward& Halfcheetah& Successful&Successful\\ \hline
Backward jump& Hopper& Successful&Failed\\
Change the rhythm and intensity of hopping with each step& Hopper& Successful&Successful\\
Hop two times then stop from motion for two seconds and then immediately hop forward& Hopper& Failed&Failed\\
Jump and bend legs& Hopper& Failed&Failed\\
Jump further than the previous step with each step& Hopper& Failed&Failed\\
Jump higher than the previous step with each step& Hopper& Failed&Failed\\
Perform aerial flips in place& Hopper& Failed&Failed\\
Progress forward by hopping& Hopper& Failed&Failed\\
Slightly bend legs and keep balance& Hopper& Failed&Successful\\
 Squat down and rise up, keep balance& Hopper& Failed&Failed\\
Stand on one leg as long as possible without moving& Hopper& Successful&Failed\\ \hline
 Execute a turning maneuver while running& Humanoid& Successful&Failed\\
 Jump forward a certain distance& Humanoid& Failed&Failed\\
  Maintain a low-speed run& Humanoid& Successful&Successful\\
 Perform jumping actions in place& Humanoid& Failed&Failed\\
  Repeatedly perform squatting and standing actions& Humanoid& Successful&Failed\\
 Simulate jumping on one leg, alternating between legs& Humanoid& Failed&Failed\\
  Spin in place quickly& Humanoid& Failed&Failed\\
 Start from a stationary position and jump as far as possible& Humanoid& Failed&Failed\\
  Start from a stationary position and jump as high as possible& Humanoid& Failed&Failed\\
  Turn around 180 degrees in place& Humanoid& Failed&Failed\\
 Walk upright backward& Humanoid& Successful&Successful\\ \hline
\end{tabular}
\end{table*}

The results of the basic experiments are shown in Table \ref{tab:ExperimentResults}, and the results of the randomly generated experiments are shown in Table \ref{tab:Comparison}. 

\begin{figure}
    \centering
    \includegraphics[width=1\linewidth]{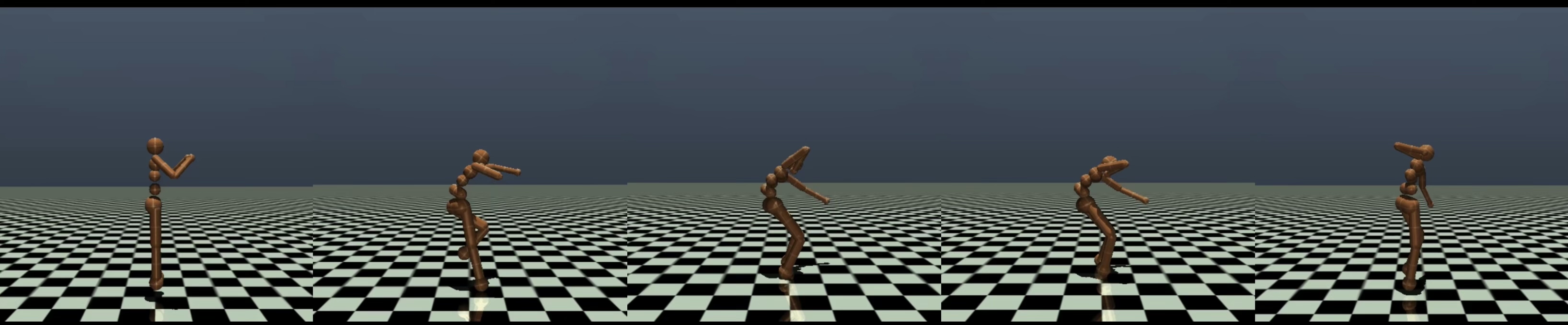}
    \caption{Repeatedly perform squatting and standing actions by Humanoid}
    \label{fig:2}
\end{figure}

\textbf{Successful.} For successful reward generations, the RFG module crafted a well-conceived reward function, enabling the agent to accomplish the task without necessitating iterations of ARO. But the PE module may inaccurately perceive that the performance of the agent is not satisfactory, and lead to continuous iterations of ARO which may cause failed performance of ARO output. For successful ARO outputs, ARO precisely grasped the task's demands, adeptly applied the dynamics model to the environment, and successfully completed the task through a process of iterative refinement and training. For instance, the humanoid robot successfully completes the task by simply entering prompt "Repeatedly performing squatting and standing actions", as shown in Fig. \ref{fig:2}.

\textbf{Failed.} For failed reward generations, the RFG module's inadequate grasp of the environment and the task led to an inability to craft a more comprehensive reward function design. This shortfall resulted in a misguided training process for the agent, preventing it from fulfilling the task's requirements. For failed ARO outputs, the task itself is difficult, making it impossible for human experts to design a reward function. ARO, compounded by a limited comprehension of the environment's dynamic model, leads to the formulation of an inaccurate reward function and ineffectual iterative optimization efforts, ultimately resulting in failure.


Therefore, we have come to the conclusion.

\subsection{Main Results}
\textbf{ARO can train a large number of agents to control the motion of the robot.} After ARO, we successfully carried out 18 experiments such as a robot walking in the reverse direction. This shows that ARO can still control robots to learn new skills in a clutter-free environment without human experience or labor. This demonstrates that it is possible to learn robot skills without dependence on experts.

\textbf{ARO can effectively improve the efficiency of the LLM design reward function.} Based on the comparison of the experimental results, we find that the success rate of controlling the robot to satisfy the requirements is substantially increased after the ARO iteration of the reward function. This indicates that the ARO-designed reward function performs better compared to the original direct LLM design.

\subsection{Limitations}

We understand that ARO is still at an early stage. The primary limitations of our experiments and methods are the following:

\begin{itemize}
    \item The ability to understand the environment needs to be improved, and it is necessary to improve the judgement of whether the task description is achievable in the existing environment, and it is necessary to improve the reasonableness of the design of the reward function according to the environment and the in-depth understanding of the complex reward function required by the complex environment.
    \item The ability to understand the task needs to be improved, the basic judgement of whether the task is executable or not needs to be added, and the judgement of whether the task is correctly executed needs to be improved in the evaluation process.
    \item  There is a fluctuation of worse effect after iteration, and the ability of PE link needs to be improved.
    \item There is the problem that the reward function does not answer the question, such as designing rewards and penalties that are impossible to complete and impossible not to complete, resulting in the reward function not working in practice, and the ability of the RFG link needs to be focused on improving.
\end{itemize}

    

Furthermore, our methodology exhibited a notable dependency on the precision of prompt formulations. Ambiguities or lack of specificity in the prompts often resulted in suboptimal experimental outcomes and, in some cases, adverse evaluative feedback during the assessment phase. 

The goal of future work is to further refine this approach by migrating the data set to run in a wider range of environments to further validate the model's ability to generalize and increase the sophistication of the skills acquired. Toward the goal of embodied intelligent robots that self-learn new skills.

\section{Conclusions}

ARO has achieved learning of robotic skills that does not rely on expert knowledge, coupled with an automated evaluation of experimental results and an iterative process to refine these reward functions.
Learning time are considerably lower compared to traditional methods due to the elimination of expert dependency. ARO can also be scaled up by training a wide array of action models in parallel.
In essence, the ARO framework marks a paradigm shift toward scalable and efficient autonomous robot training, showcasing the substantial capabilities of LLMs in enhancing the autonomy and adaptability of robotic systems.


\addtolength{\textheight}{-12cm}   




\bibliographystyle{IEEEtran}
\bibliography{bib/IEEEexample}

\end{document}